\documentclass[fleqn,10pt]{wlscirep}
\usepackage[utf8]{inputenc}
\usepackage[T1]{fontenc}
\usepackage{dirtytalk}
\title{How should the advent of large language models affect the practice of science?}

\author[a,q,1,*]{Marcel Binz}
\author[b,1]{Stephan Alaniz}
\author[d]{Adina Roskies}
\author[e]{Balazs Aczel}
\author[c]{Carl T.\ Bergstrom}
\author[f]{Colin Allen}
\author[g]{Daniel Schad}
\author[h,i]{Dirk Wulff}
\author[c]{Jevin D. West}
\author[j]{Qiong Zhang}
\author[k]{Richard M. Shiffrin}
\author[l]{Samuel J.\ Gershman}
\author[m]{Ven Popov}
\author[c,2]{Emily M.\ Bender}
\author[n,2]{Marco Marelli}
\author[o,p,2]{Matthew M.\ Botvinick}
\author[b,3]{Zeynep Akata}
\author[a,q,2,3]{Eric Schulz}

\affil[a]{Max Planck Institute for Biological Cybernetics}
\affil[b]{University of Tübingen}
\affil[c]{University of Washington}
\affil[d]{University of California Santa Barbara}
\affil[e]{Eötvös Loránd University, Budapest}
\affil[f]{University of California Santa Barbara}
\affil[g]{Health and Medical University, Potsdam}
\affil[h]{Max-Planck-Institute for Human Development}
\affil[i]{University of Basel}
\affil[j]{Rutgers University, New Brunswick}
\affil[k]{Indiana University, Bloomington}
\affil[l]{Harvard University}
\affil[m]{University of Zurich}
\affil[n]{University of Milano-Bicocca}
\affil[o]{Google DeepMind}
\affil[p]{University College London}
\affil[q]{Helmholtz Center for Computational Health, Munich}
\affil[1]{Joint first authors}
\affil[2]{Perspective leaders}
\affil[3]{Joint senior authors}
\affil[*]{marcel.binz@tuebingen.mpg.de}


\begin{abstract}
Large language models (LLMs) are being increasingly incorporated into scientific workflows. However, we have yet to fully grasp the implications of this integration. How should the advent of large language models affect the practice of science? For this opinion piece, we have invited four diverse groups of scientists to reflect on this query, sharing their perspectives and engaging in debate.
Schulz et al. make the argument that working with LLMs is not fundamentally different from working with human collaborators, while Bender et al. argue that LLMs are often misused and over-hyped, and that their limitations warrant a focus on more specialized, easily interpretable tools. Marelli et al. emphasize the importance of transparent attribution and responsible use of LLMs. Finally, Botvinick and Gershman advocate that humans should retain responsibility for determining the scientific roadmap.
To facilitate the discussion, the four perspectives are complemented with a response from each group. By putting these different perspectives in conversation, we aim to bring attention to important considerations within the academic community regarding the adoption of LLMs and their impact on both current and future scientific practices. \\

\noindent\textbf{Significance Statement:} \\Artificial intelligence (AI) is reshaping the way researchers conduct science. Large language models (LLMs) in particular have received attention for their apparent versatility in reading, analyzing, and writing text. Yet, at the same time, there are also considerable concerns that come with the use of this technology, which could potentially place our scientific integrity at risk. This raises the question: how should LLMs affect the practice of science? In this manuscript, we present four perspectives on this issue, each one from a different group of researchers. The perspectives demonstrate the extent of both issues and opportunities, ranging from encouraging the use of LLMs in science to a stern warning against using LLMs in most, if not all, of the proposed cases.

\end{abstract}
\begin{document}

\flushbottom
\maketitle

\thispagestyle{empty}

\newpage
Language models are statistical models of human language that can be used to predict the next token (e.g., a word or character) for a given text sequence. Even though these models have been around for decades \cite{bengio2000neural, Jurafsky2009}, they have recently experienced an unprecedented renaissance: by training enormous neural networks with billions of parameters on data sets with trillions of tokens, researchers have observed the emergence of models whose abilities can go beyond mere text generation and conversational skills \cite{brown2020language}.

Modern \emph{large} language models (LLMs) are, amongst other things, able to solve selected university-level math problems \cite{drori2022neural}, support language translation \cite{kocmi2023large}, or answer questions in a bar exam with high accuracy \cite{katz2023gpt}, out of the box and without additional training. Given the range of these capabilities, it seems possible that these systems will have an enormous impact on our society, leaving their mark on the labor market \cite{eloundou2023gpts}, the education system \cite{kasneci2023chatgpt}, and many other parts of our daily lives.

We\,---\,as scientists\,---\,may therefore wonder: how will the advent of LLMs affect the practice of science? Finding answers to this question is urgent as LLMs are already starting to permeate the academic landscape \cite{peres2023chatgpt, lund2023chatting, hill2023chat, zheng2023chatgpt, lund2023chatgpt, transformer2022can, birhane2023science, fecher2023friend, stokelwalker2023chatgpt}. For instance, in 2022, MetaAI released the first science-specific LLM (under the name Galactica) aimed to support researchers in the process of knowledge discovery \cite{taylor2022galactica}. Even more recently, Terence Tao, a Fields Medal-winning mathematician, proclaimed \cite{tao2023} that \say{the 2023-level AI can already generate [$\ldots$] promising leads to a working mathematician [$\ldots$]. When integrated with tools such as formal proof verifiers, internet search, and symbolic math packages, I expect, say, 2026-level AI [$\ldots$]  will be a trustworthy co-author [$\ldots$].} 

Yet while there have been claims of immense potential for this technology for the advancement of science, there are also considerable concerns that need to be taken into account. For instance, the aforementioned Galactica model had to be taken offline after just three days because it was heavily criticized by researchers for fabricating information, such as \say{fake papers (sometimes attributing them to real authors), and [...] wiki articles about the history of bears in space} \cite{heaven2022why}. Furthermore, even though LLMs often achieve state-of-the-art performance on existing benchmarks, it remains debated whether this reflects genuine understanding, or whether they are merely regurgitating the training set, thereby acting like stochastic parrots \cite{bender2021dangers}. It has been, for instance, repeatedly demonstrated that even the most capable models at present fail at basic arithmetic problems such as multiplying two four-digit numbers \cite{arkoudas2023gpt}. Flaws like these are especially concerning if we intend to utilize LLMs for research purposes, and could endanger the integrity of science if we act carelessly.

The objective of the present article is to provide researchers with different opinions and a forum to voice and discuss their perspectives on if and how we \emph{should} make use of LLMs in the context of science. To facilitate this discussion, we will first highlight a few applications where LLMs have the potential to positively impact science, followed by pointing out some of the issues that come with them.

\section*{Background: applications of LLMs in science}

LLMs find their most obvious use case as a supporting tool for scientific writing. For example, as proofreaders of manuscript drafts, they can aid in rectifying grammatical errors, improving the writing style, and ensuring adherence to editorial guidelines. Beyond manuscript composition, LLMs could prove valuable for data acquisition and analysis in domains that were traditionally reliant on manual human labor~\cite{gilardi2023chatgpt,wulff2023automated}. Researchers have even suggested using LLMs as potential substitutes for human participants, as proxies~\cite{dillion2023can} or for pilot studies~\cite{hutson2023pigbots}. In computational fields, LLMs could speed up prototyping by writing code~\cite{roziere2023codellama}, while a human-in-the-loop would guide these processes, correct LLM-generated errors and ultimately decide which ideas warrant further pursuit.
Moreover, researchers might experiment with employing LLMs at certain stages of research with progressively reduced supervision~\cite{sanmarchi2023step}, potentially leading to increased automation in some aspects of scientific exploration and discovery.

While the potential influence of LLMs on the practice of science is immense, there are pressing issues that come with the use of LLMs in the context of science.
For instance, when an LLM helps us to write text, who ensures that its output is not subject to plagiarism issues~\cite{dehouche2021plagiarism}? LLMs learn from web-sourced text data, acquiring inherent biases~\cite{liang2021towards, coda2023inducing, hutchinson2020socialbias} and\,---\,in some cases\,---\,replicate excerpts from their training data~\cite{carlini2021extracting}. When an LLM is used for data analysis, what happens when it hallucinates data? The content generated by LLMs can contain errors or fabricated information, presenting a potential threat to the integrity of scientific publishing~\cite{zheng2023chatgpt}. When an LLM suggests an idea, who gets credit for it? The general consensus within the scientific community seems to indicate that LLMs are not eligible for (co-)authorship \cite{king2023place} as they cannot be held accountable for upholding scientific precision and integrity. Leading AI conferences such as ICML\footnote{\url{https://icml.cc/Conferences/2023/llm-policy}} and ACL\footnote{\url{https://2023.aclweb.org/blog/ACL-2023-policy/}}\,---\,as well as journals such as Science\footnote{\url{https://www.science.org/content/page/science-journals-editorial-policies}}, Nature\footnote{\url{https://www.nature.com/nature-portfolio/editorial-policies/ai}} and PNAS\footnote{\url{https://www.pnas.org/author-center/editorial-and-journal-policies\#authorship-and-contributions}}\,---\,have already adopted policies to limit the involvement of LLMs. However, it remains an open question how strong these regulations should be and if and how the usage of LLMs should be acknowledged.

These\,---\,and many other\,---\,issues raise the questions: How \emph{should} the advent of LLMs affect the practice of science? Do LLMs actually improve our scientific output or are they rather hindering good scientific practice? To what extent should they be used given the ethical and legal issues that come with them? We believe these to be highly non-trivial questions without an obvious answer and have therefore invited four groups of researchers to provide their perspectives on them. These perspectives were selected to cover a broad spectrum of opinions in order to spark a constructive discussion. Each of the perspectives is accompanied by a response from each group. We conclude this article with a short general discussion in which we attempt to identify common themes.

\section*{Perspective -- LLMs: more like a human collaborator than a software tool}

\noindent \textbf{Contributors}: Eric Schulz, Daniel Schad, Marcel Binz, Stephan Alaniz, Ven Popov, and Zeynep Akata \\

Most researchers in our labs already frequently employ LLMs in their everyday work. They use them, amongst other things, to finetune and revise their drafts, as a supporting tool for programming, to suggest formulations for research items such as questionnaires or experimental instructions, and to summarise research papers. We have observed a significant increase in quality in all of these areas after the widespread adoption of these models. While our personal experience may be biased, there are several studies supporting the idea that LLMs can facilitate writing \cite{herbold2023ai}, coding \cite{poldrack2023ai}, and knowledge extraction \cite{goyal2022news}. In the future, we expect these models to be even more deeply integrated into the scientific process, taking on roles similar to a collaborator with whom one can develop and discuss ideas.

Indeed, we believe that working with LLMs will not be fundamentally different from working with other collaborators, such as research assistants or doctoral students. LLMs are not perfect and have limitations and biases that could affect their performance and output. However, humans are also subject to some of the same flaws, such as errors, plagiarism, fabrication, or discrimination. If we take this perspective, it seems appropriate to view current LLMs less as traditional software tools and more as knowledgeable research assistants: they can do phenomenal work but we need to be aware that they can make mistakes.

\subsection*{Protecting the past}

It is our chief responsibility to ensure the quality and integrity of our work. There are already rules and norms about scientific practice in place to ensure this, and many of them also apply to LLMs. For instance, we should always check the accuracy and validity of the information and data we obtain, no matter the source, as well as correctly cite the sources and methods we use. That means that we should not blindly trust or rely on LLMs, but rather use them as a complement to our own expertise and judgment. Furthermore, our work can only be criticized appropriately if all information about its methodology is transparently communicated. We should therefore acknowledge the contributions of LLMs to our research, just as we would do for any other tool. Ultimately, it is\,---\,and will remain\,---\,the authors' responsibility to ensure that the appropriate scientific standards are followed, regardless of whether we use LLMs or not. 

Ensuring that our research is reproducible is one of the cornerstones of modern science. However, as many LLMs are proprietary, working with them poses a threat to this ideal. Nobody guarantees that OpenAI, Google, or other providers will not make changes to their models (in the worst case, without informing the user). In fact, this happened to us during the revision process of one of our papers, where, at some point, we could not reproduce our initial results, likely due to changes on the provider side. How should we deal with such cases? We believe that the obvious solution to this problem is to rely on open-source models where one has full control over all aspects of the model. Following a recent call for action to the European Parliament \cite{laion2023}, we therefore strongly advocate for the development of such models, such that they can become the primary tool for scientific inquiry.

\subsection*{Welcoming the future}

Paper reviewing is another area where LLMs could improve our scientific pipeline. In a recent study, Liang and colleagues \cite{liang2023can} demonstrated this potential by systematically comparing LLM-generated reviews to reviews written by human researchers. They found that \say{more than half (57.4\%) of the users found GPT-4 generated feedback helpful or very helpful and 82.4\% found it more beneficial than feedback from at least some human reviewers.} Not only does this result allow scientists\,---\,especially early career researchers\,---\,to receive high-quality, instantaneous feedback (similar to that one could get from a critical colleague with an unlimited amount of time) but it also has implications for the peer review process. Yet, the use of LLMs in the peer review process also presents one major legal obstacle: manuscripts under review are typically confidential, and hence should not be entered into proprietary LLMs. To prevent such breaches of confidentiality, the National Institutes of Health (NIH) and other institutions have rules in place that prohibit the use of LLMs for peer review.\footnote{\url{https://grants.nih.gov/grants/guide/notice-files/NOT-OD-23-149.html}} Locally hosted, open-source models are again a solution to this issue, as they provide control about which information is shared with external sources and which is not.

We also would like to point out that LLMs are a moving target, constantly evolving and becoming more capable and autonomous. This may raise new challenges and questions for the scientific community in the future, such as how to evaluate, interpret, and communicate the results generated by LLMs, or how to ensure their transparency and accountability. We welcome these challenges as an opportunity to advance our understanding and methods of science. We also encourage researchers to collaborate with each other and with LLM developers to address these issues and ensure that LLMs improve at frequently criticised skills such as providing truthful sources or acknowledging ignorance.

\subsection*{Conclusion}

In conclusion, LLMs are a valuable asset for science and should be embraced rather than feared or restricted. It becomes apparent that they are not infallible machines once we start thinking about them as knowledgeable research assistants instead of traditional software tools. Furthermore, since rules for good scientific practice are already in place, and since it is the authors' obligation to take responsibility for adhering to these rules, there is no need for novel rules with the use of LLMs. We believe that strengthening the development of open-source alternatives should be one of our top priorities, as they \say{offer enhanced security, explainability, and robustness due to their transparency and the vast community oversight} \cite{laion2023}. Finally, being conscious about the current limitations of LLMs and embracing them, will allow us to grow with the technology as LLM research finds remedies and develops complementary tools. We hope that by adopting this liberal perspective, we can foster a positive and fruitful relationship between humans and LLMs in science. Please note that the first draft of our perspective was written by an LLM (GPT-4) based on our meeting notes.

\section*{Perspective -- Science is a social process that cannot be auto-completed}

\noindent \textbf{Contributors}: Emily M.\ Bender, Carl T. Bergstrom, and Jevin D. West \\

When deciding whether to use an LLM, it is important to recognize that LLMs are simply models of word form distributions extracted from text\,---\,not models of the {\em information} that people might get from reading that text \cite{Ben:Kol:20}. Originally, such systems were used to rank or classify text. In automatic transcription, for example, an acoustic model provides a set of possibilities and the language model helps determine the most likely next word \cite{Wan:Wan:Lv:19}. Today, however, LLMs are vaunted for their ability to extrude synthetic text by repeatedly selecting a next likely token.

Trained on sufficiently large datasets and with sufficiently well-tuned architectures and training processes, LLMs  appear to produce coherent text on just about any topic, including scientific ones. Moreover, we can't help but make sense of their textual output because our linguistic processing capabilities are instinctual and reflexive \cite{Har:Moo:17}.

Proponents argue that LLMs are useful in three domains: 1) navigating science, by searching and synthesizing published literature, 2) doing science, in the sense of designing or conducting experiments or generating data, and 3) communicating science, by drafting text for publication. While certain machine approaches may be useful in each, LLMs are unlikely to outperform alternative technologies. Furthermore, they have the potential to cause downstream harms to science if their use is widely embraced.

\subsection*{Navigating science} Natural language processing has proven useful in sorting through an ever-growing body of scientific literature. Information retrieval and extraction techniques, as implemented in academic search engines (e.g. ref.~\cite{Kin:Ana:Aut:23}), have helped researchers discover relevant prior work. 
Will LLMs supplant other NLP approaches? We doubt it. The inappropriateness of LLMs as text generators and synthesis machines was highlighted in Meta's Galactica debacle. That system---taken off-line after three days in response to intense criticism for its abysmal performance---had been trained on scientific text and promoted as a tool to ``summarize academic papers, solve math problems, generate Wiki articles, write scientific code, annotate molecules and proteins, and more.''\ \cite{heaven2022why}.
But training an LLM on scientific papers doesn't guarantee that it will output scientifically accurate information. As Meta discovered, the use of LLMs yields text ungrounded in any communicative intent or accountability for accuracy.

One might hope that LLMs could at least be used to summarize a set of papers. Extractive summarization systems \cite{narayan2018ranking} already do this; will LLMs perform better?  Will people tend to over-rely on system output rather than using it as a starting point? What are the costs of false negatives, i.e., important points not included in the generated summary? How will errors generated by LLMs, which then become training data for future LLMs, get amplified? 

\subsection*{Doing science}  LLMs are just one of many technologies 
dubbed ``artificial intelligence'', but their surprising capacity to perform what amounts to a fancy parlor trick has drawn outsized attention. That’s a mistake. LLMs may be adequate for specific linguistic tasks such as grammar checking, automatic transcription, and machine translation (including code generation), but they are unlikely to provide an effective basis for most tasks involved in hybrid human-machine science. Even where they do appear to be moderately effective, they are known to be brittle to input variation \cite{hodel2023response}. The future of machine-aided science will not be a massive, one-size-fits-all, universal application of LLMs, but rather an ensemble of bespoke and often lightweight models that have been designed explicitly to solve the specific tasks at hand\,---\,and, crucially, evaluated in terms of those specific tasks. Such approaches also have a major advantage where interpretability is concerned. If researchers want to understand output variation, let alone find ways to fine-tune the architecture to generate better results, they need to steer away from technologies as opaque as LLMs. But instead, the ongoing hype around LLMs is drawing funding and brainpower away from more promising, targeted approaches.

Not only are LLMs being explored as aides to researchers; numerous proposals suggest that they can stand in for test subjects \cite{törnberg2023simulating}, survey participants \cite{arg:us:ful:2023}, or data annotators \cite{Gil:Ali:Kub:23}. Such arguments derive from a failure to understand that LLMs model the output of sequential linguistic tokens, not concepts, meanings, or communicative intent. If we are looking to study the opinions or behavior of human beings, we need to work with actual people. 

\subsection*{Communicating science} By design LLMs generate form without substance. The synthetic text that systems output constitutes neither ideas, nor data\,---\,and it certainly is not a reliable information source. This notion of generating statements that no one intended is anathema to the spirit of scientific inquiry. Automatically generating something that looks like a manuscript is very different from the iterative process of actually writing a manuscript. Yet the output can be difficult to distinguish, particularly in a cursory read or by inexpert readers. Some proponents argue that LLMs can relieve scientists of the drudgery of writing papers and free them up to get on with the serious business of ``doing science'' \cite{conroy2023chatgpt}. This false dichotomy between communication and investigation reflects a fundamental misunderstanding of the nature of science \cite{latour2013laboratory}
that devalues the communicative aspects of science and ignores the role of writing in the process of formulating, organizing, and refining ideas.

Downstream, LLMs threaten the notion of scientific expertise, shift incentive structures \cite{partha1994toward}, and undermine trust in the literature. When human authors review or acknowledge prior literature, we are assured that they are  familiar with the field and striving to situate their results therein. If LLMs write our introductions, we lose these guarantees. Moreover, notions of systematic review are undercut by the randomness inherent in LLM output. Finally, when a LLM generates a literature review, the claims that it generates are not directly derived from the manuscripts it cites.  Rather, the machine creates textual claims, and then predicts the citations that might be associated with them. Obviously this practice violates all norms of scholarly citation. At best, LLMs gesticulate towards the shoulders of giants. 

Driven by quantitative metrics and the strong incentive to publish, researchers may opt to trade off quality for speed by letting LLMs do much of their writing. Widespread use of one or a few LLMs could undercut epistemic diversity in science. When asked to provide a hypothesis, experiment, or mode of explanation, LLMs may repeatedly offer similar solutions, instead of leveraging the parallel creativity of an entire science community. 

Worse still, opportunistic or malicious actors could use LLMs to generate nonsense at scale with minimal cost. (This is not an argument against using LLMs appropriately, but we need to be prepared for such behavior). Lazy authors could boost their publication counts by shotgunning machine-generated papers to low-quality journals. Predatory publishers could feign peer review using LLM output. Bad actors could overwhelm the manuscript submission system of a target journal (or even a target field) with a massive volume of fake papers. Or an investigator's work could be targeted with a deluge of spurious machine-generated critiques on post-publication peer review platforms such as Pubpeer. 

Finally, LLMs may cause considerable collateral damage to science education. For example, as LLMs slash the cost of generating seemingly authoritative text, the web will be flooded with low-quality, mistake-ridden tutorials designed to capture advertising revenue. At present, search engines' ability to discriminate is more or less the only line of defense. That's worrisome.

\subsection*{Conclusion}

In conclusion, LLMs are often mis-characterized, misused, and over-hyped, yet they will certainly impact the way we do science, from search to experimental design to writing. The norms that we establish now around their use will determine the consequences far into the future. We should proceed with caution\,---\,and evaluate at every step.

\section*{Perspective -- LLMs in scientific practice: a matter of principles, not just regulations}

\noindent \textbf{Contributors}: Marco Marelli, Adina Roskies, Balazs Aczel, Colin Allen, Dirk Wulff, Qiong Zhang, and Richard M. Shiffrin \\

A moderate perspective on the potential impact of LLMs on scientific practice holds that, while it is important to be mindful of the dangers, their application seems largely beneficial, insofar they offer a much needed support in day-to-day research activity and may alleviate major obstacles to scientific advancement. This is evident when LLMs are applied as editing tools: they provide a writing aid that leaves researchers with more time for brainstorming ideas and analysis, may help mitigate disparities between different scientific communities, and remedy some of the disadvantages for researchers who are not native speakers of English \cite{amano2023manifold}. Also, LLMs can access a broader range of literature than any individual researcher could, potentially offering valuable support for literature analysis and hypothesis generation \cite{jumper2021highly}, with a reach that goes beyond one’s research specialization. 

However, although any new technology may be used for good or evil,  some technologies afford opportunities for good or evil uses more than others. LLMs have disruptive potential that is ever more evident, and such disruption must be kept at bay if the goal is to prevent “evil drifts”. One perspective might hold that strict regulation is required, but regulation carries with it many costs that might be best avoided if it is kept moderate. A preferable approach may be to adopt clear principles guiding the way this technology should be used, principles that cannot just focus on efficiency and overall utility. Such principles include transparency, accountability, and fairness.

\subsection*{A matter of transparency} In science, transparency is of indispensable value. When used as writing tools, researchers must acknowledge the reliance on LLMs so that readers are on notice that the text is (at least partially) AI-generated. Authors should make explicit which LLMs were applied and how, as part of the method sections or in a separate dedicated statement. This could be achieved by relying on already existing solutions; for example, the CrediT taxonomy\footnote{\url{https://credit.niso.org/}} could also be used to code the nature of AI contribution, even if AI is not to be recognized as a coauthor. Ideally, in the spirit of open science, authors shall publicly release their prompts along with the corresponding LLM responses as supplementary materials, and reference such archives in the manuscript. Importantly, transparency does not only pertain to the way we exploit LLMs, but to the systems themselves. LLMs are not, strictly speaking, anything new. Models that are analogous to current LLMs in structure, spirit, and basic mechanisms have been part of the scientific debate for decades \cite{gunther2019vector}. However, such older models were unambiguous about their architecture and training, if not openly released. Current LLMs are often not held to the same scientific standards as their ancestors, being widely applied even when their inner workings and training data remain undisclosed. This causes substantial issues in estimating the actual performance of such models (and, importantly, the possibility of data contamination \cite{golchin2023time}). As a scientific community valuing greater transparency, we should favour the systems that are taking some steps in that direction \cite{li2023starcoder}.

\subsection*{A matter of accountability} It must be acknowledged that LLMs are instruments of human agency, and researchers should be held accountable for any scientific product they present to the community, irrespective of the extent to which this was obtained through the application of automatic systems. The Association for the Advancement in Artificial Intelligence has released clear guidelines in this respect: “Attribution of authorship carries with it accountability for the work, which cannot be effectively applied to AI systems $\ldots$ Ultimately, all authors are responsible for the entire content of their papers, including text, figures, references, and appendices.” For example, LLMs are known to “hallucinate” and produce factually incorrect responses \cite{walters2023fabrication}. They can fabricate bibliographic citations, omit important references when summarizing literature, and potentially plagiarize text written by another researcher. The burden to verify that LLM-produced texts are accurate and that LLM-proofread texts are consistent with the original message remains with the individual authors. Similarly, LLMs can be particularly poor at logic and deductive reasoning \cite{kocon2023chatgpt, liu2023evaluating}, so using them for analysis may lead to false conclusions. The onus is on the user to make sure that what LLMs produce is worth pursuing. Researchers must hence have strategies for assessment over AI-related content; a good practice would be to have clear quality criteria and verification methods defined before using LLMs. Scientists should not underestimate the time and effort that such vetting will take, and should weigh the efficiency of LLM application against these costs.

\subsection*{A matter of fairness} AI in general and LLMs in particular have the potential to deeply affect us at a societal level. Science, as any human endeavour, is not immune to this. As a community we must make all possible efforts to guarantee that reliance on LMMs does not violate basic fairness principles. Indeed, current language models reflect mostly WEIRD (Western Educated Industrialized Rich Democratic) populations and cannot easily be prompted to represent non-WEIRD communities \cite{durmus2023towards, atari2023humans}. This leads to biases in writing and annotation, potentially reinforcing distortions in citations and marginalization of already marginalized scientists. Moreover, it may have negative consequences in terms of equitable research, given that LLMs are also more accessible to WEIRD populations. More generally, LLMs will, for known or unknown reasons, favour some perspectives or sources over others \cite{santurkar2023whose}. These systematic patterns must be recognized and taken into account, to avoid unprincipled biases affecting the direction of research and possibly the relative success of careers. 

\subsection*{Conclusion} The impact that LLMs are having on scientific practice cannot be understated. Given the current trend, at the time you are reading these words such impact will likely be much larger than it is as we write this piece. Precisely how LLMs will influence the practice of science in the future cannot be entirely predicted and countering such a revolution with strict, preconceived norms is a losing battle. Rather, establishing principles and shared values in the scientific community constitutes the ideal foundation when deciding how to manage these rapidly changing technologies. Most importantly, we need to train students and each other to build upon such principles in order to become appropriately skeptical towards these systems and their outputs.

\section*{Perspective -- AI can help, but science is for people}

\noindent \textbf{Contributors}: Matthew M. Botvinick and Samuel J. Gershman \\

Like many forms of technology, AI can substitute for human labor. With the advent of LLMs, the relevant kinds of labor begin to overlap with high-level human cognitive work, including the activities involved in science \cite{birhane2023science}. As LLMs improve, their ability to substitute for human scientific labor will be a major boon. However, we argue here that two core aspects of scientific work should be reserved to human scientists. 

\subsection*{AI and scientific labor}

Over time, the labor involved in scientific research has become progressively more onerous, sometimes now bordering on the intractable. Assimilating current knowledge has become more difficult in the face of increasingly voluminous literatures. Generating new questions, hypotheses and experimental tests has become more challenging, as the search problem entailed by each has become more complex. Drawing conclusions from experimental results has become harder as the size and complexity of datasets has exploded. And communicating and debating scientific conclusions has become more challenging for reasons including an overtaxing of peer review systems \cite{flaherty2023}. Given the increasing costs of scientific labor on these fronts, it’s no surprise that progress across multiple scientific fields appears to have slowed \cite{park2023papers}. 

In the long run, AI may help us cope with the increasing demands of scientific work. Through the kinds of application detailed in the introductory essay above, AI may help us scale up, by making each step in the research cycle cheaper. In some cases, AI may eventually perform some forms of scientific labor better than human scientists, including the work of generating new hypotheses \cite{davies2021advancing}. Even in present-day forms, AI may be useful on some fronts, as reviewed in the introduction. Of course, as widely discussed, current systems are too unreliable to deploy without caution and oversight (see accompanying commentaries), and only time will tell how feasible it may be to overcome current limitations. 

However, in addition to addressing present-day shortcomings, it’s equally important to look into the future and consider what kind of AI tools we actually want for science in the long run. Given that AI can be applied to all phases of scientific work, one aim might be to build a full-fledged AI scientist, one that can do everything a human scientist now does: a full-spectrum replacement for human scientists. To us, this prospect is deeply unappealing. Why? Because there are particular aspects of science that we simply would not want to delegate to AI, even in a scenario where technical limitations presented no barrier. In particular, there are two core aspects of science that should be left to people. As we now explain, one of these is normative and the other epistemic.  

\subsection*{The normative aspect of science}

Any scientific discipline must continually ask, What problems shall we work on? How this question gets answered, both within individual labs and across whole research communities, is a complex affair, but it centers on judgments concerning the ‘interest’ and ‘significance’ of candidate problems, as well as their ‘timeliness,’ including their amenability to study under prevailing material and ethical constraints. Such judgments are informed by hard data; we obviously cannot reduce them to purely social constructions. However, at the same time, judgments of interestingness, significance and timeliness are inherently tied to culturally and historically grounded sensibilities and mores. This is not a corruption or impurity in scientific thought and procedure. Cultural sensibilities and patterns of thought are fundamental to scientific prioritization. 

This point will be especially salient to students of the history of science, because the sensibilities and mores that inform science evolve over time. Just as scientific theory changes over the years, so do the ethical commitments and intellectual priorities that underlie science. This is evident in the fact that we no longer approach homosexuality as a disorder, or study genetics through the lens of eugenics. It shows in growing restrictions on animal experimentation. And it shows in the attention that Western climatologists now pay to regions historically neglected. 

We argue that the normative aspect of science should not be ceded to AI systems, no matter how capable those systems become. People should stay in the driver’s seat, determining the direction of travel for science. Certainly, AI systems may be helpful partners in deliberation, especially as techniques for AI value alignment improve \cite{gabriel2021challenge}. However, aligning a system to currently prevailing human views is different from allowing that system to govern the evolution of human views. In science, the ultimate driving force in that evolution should remain human. We are the moral agents in the room, and we shouldn’t forget it.   

\subsection*{The epistemic aspect of science}

Obviously, a central goal of basic science is understanding the natural world. If we are going to do science with AI tools, the question arises: ‘whose’ understanding matters? Would it be satisfactory to have AI systems that in some sense understand aspects of nature, but which don’t make that understanding accessible to people? From an engineering standpoint that might be fine. However, if it’s basic science we’re talking about, we shouldn’t let go of the core objective, which is not just practical but epistemic. We cannot cede understanding to artificial systems. We should insist on human understanding remaining a core goal of science. 

Of course, it may be that because of limitations on human cognition, AI systems may someday be able to represent some aspects of nature that we cannot, just as existing AI systems master aspects of complex board games that elude even highly skilled human players \cite{silver2017mastering}. Even in these cases, however, we should strive to extract as much human insight from AI systems as possible \cite{lemos2023rediscovering}. We shouldn’t lose track of what basic science is for. 

\subsection*{Conclusion}

AI promises to deliver great value in science, just as in many other domains. We believe its potential should be embraced. However, at the same time that we strive to break through the current limitations of AI to access its benefits, we should also think through our long-term goals in developing this technology. In the end, the two areas of science we’ve proposed to protect — one normative, the other epistemic —  are two reflections of a more general bound on AI’s proper domain. We might call this the subjective limit. Unlike AI systems, people have a ‘point of view,’ which cannot be automated because it’s inherently subjective \cite{botvinick2023}. This point of view includes knowledge that is meaningful to us (the epistemic view) and values that are meaningful to us (the normative view).  Machines might have their own knowledge or values, and these might be aligned with ours, but the alignment problem is fundamentally yoked to our subjective views. This principle applies in science, as in all human-centered activities.

\section*{Response \textmd{by Eric Schulz, Daniel Schad, Marcel Binz, Stephan Alaniz, Ven Popov, and Zeynep Akata}}

We have argued that one should think of working with LLMs less as using a traditional software tool and more as working with a human collaborator and that this perspective allows us to better understand their shortcomings. This view actually resonates with many of the points raised in the other perspectives. For example, Marelli et al. write that \say{we should not blindly trust or rely on LLMs, but rather use them as a complement to our own expertise and judgment}, and Bender et al. argue that collaboration in science means iterating over outputs many times. Like working with a human collaborator, working with LLMs is an iterative process in which we constantly check for facts and logical consistency, revise arguments, and identify new connections. This process takes time and is more than just booting up an LLM and copy-pasting its outputs; as nicely put by Marelli et al., we \say{should not underestimate the time and effort that such vetting will take, and should weigh the efficiency of LLM application against these costs.}

We would like to stress that the notion that \say{LLMs are simply models of word form distributions extracted from text} oversimplifies both their capabilities and the additional engineering effort involved in modern LLMs. If one takes steps like reinforcement learning from human feedback \cite{stiennon2020learning} or instruction tuning \cite{10.5555/3618408.3619349} out of the equation, the outputs produced by such models are rather uninspiring (anyone who has ever worked with a plain LLM can attest to this). However, with those ingredients, LLMs do not just mimic language patterns; they can also synthesize concepts, critically evaluate their own outputs, and assist in problem-solving by processing vast amounts of data. 

Bender et al. argue that \say{the future of machine-aided science will not be a massive, one-size-fits-all, universal application of LLMs, but rather an ensemble of bespoke and often lightweight models that have been designed explicitly to solve the specific tasks at hand [...].} We believe LLMs are widely adopted precisely because they are a universal tool to accomplish many tasks. Not only does that remove the need to build specialized tools for each application, but it also eradicates the time it takes to learn them. Like human collaborators, who bring a diverse range of skills to a project, LLMs offer a breadth of knowledge that can be tailored to specific needs, e.g., as shown with the finetuning of coding LLMs \cite{roziere2023codellama}. There are -- of course\,---\,applications that benefit from purposefully designed tools, but we believe that the percentage of such applications is modest once we take the time required to develop and learn such tools into account.

Finally, there is the question of how much autonomy we want to transfer to LLMs or other AI systems. Botvinick and Gershman advocated that people should retain control over certain aspects of the scientific pipeline, such as deciding which topics to work on. We do not think that such a constraint is necessary. For example, if in the future, an LLM (or any other AI system) decides to work on a topic that it deems interesting, and this LLM has proven itself to select topics in a very fruitful and productive manner, should we stop it? We do not think so as long as ethical and legal guidelines are followed. Deciding on scientific topics is hard, and it is often not a priori known which research directions will be fruitful. Therefore, we should take any help we can get. Human researchers and AI systems bring complementary strength to the table, and acknowledging this collaborative spirit enables us to leverage the best out of both worlds.

\section*{Response \textmd{by Emily M.\ Bender, Carl T. Bergstrom, and Jevin D. West}}

Science is a social process. It cannot be auto-completed. Its agents\,---\,real scientists\,---\,are as much the product of this process as the results recorded in papers. 

LLM optimists envision a new world, where machines write, review, and even do much of the science. Even the less extreme narrative wherein LLMs simply aid researchers suffers from a misplaced and almost Taylorist \cite{taylor_principles_1913} optimism regarding production efficiency. Science is not a factory, churning out widgets or statistical analyses wrapped in text. For a factory, producing one more car per day is progress. For science, the goals are to understand our world\,---\,not to produce more artifacts that look like scientific papers. If science were a paper factory, we too would indulge in LLM euphoria and might even claim a significant resulting improvement in quality coming out of our labs. But we cannot equate papers and progress. Papers are but messages that we send one another to coordinate our collective quest for scientific understanding.

We don't, however, believe that any new mandates are required prohibiting the use of LLMs. All ill-advised use cases are already contrary to the norms of science: Using LLMs as stand-ins for human subjects or annotators amounts to fabricating data; using LLMs to write first drafts runs afoul of prohibitions against plagiarism, as it is impossible to discern the source of any string produced by an LLM; treating LLMs as co-authors contravenes norms around authorship, since LLMs are not the sort of thing that can be accountable for paper contents; using LLMs to produce peer reviews is tantamount to abrogating our responsibility to deeply evaluate the methods, reasoning and conclusions of our peers' work.

When contemplating how LLMs will affect science, we should not underestimate the temptation to use them under deadline pressure or in response to publish-or-perish threats to job security. Nor should we underestimate the time needed to fact-check all LLM output---not only for the inevitable and frequent errors but also to assess whether citations are accurate. We note that there are no published user studies that quantify just how much effort this checking process is, nor how accurately researchers can carry it out, especially while working under pressure.
Norms of plagiarism and the weight of reputation will hopefully counterbalance the unfettered use of this new technology. 

To reason appropriately about when LLMs are suitable within science, it is critical to avoid anthropomorphizing them. These models aren't research assistants. They are tools. They don't make mistakes like junior (or senior!)\ researchers do: People can take responsibility for, and learn from, mistakes. Tools produce errors; thus people using the tools have a responsibility to understand their affordances and use them with care. 

Similarly, understanding LLMs as tools positions us to ask: Is this the best tool for this task? Often, we expect, LLMs are not. Even setting aside the closed proprietary models, their attendant failures of transparency, and the stochastic nature of LLM output, we expect that bespoke models designed for specific tasks will be more efficient, performant, interpretable, and easier to fix when not functioning well.

Ultimately, science is a conversation and the interlocutors are the scientists. Synthetic text-extruding machines, designed only to produce plausible-sounding prose, are not fit participants in that conversation and should not be treated as such.

\section*{Response \textmd{by Marco Marelli, Adina Roskies, Balazs Aczel, Colin Allen, Dirk Wulff, Qiong Zhang, and Richard M. Shiffrin}}

In our proposal concerning the application of LLMs in science, we aimed for a moderate perspective. In that spirit, we think that such systems can be profitably incorporated into  scientific practice (in line with Schulz et al.), but we also recognize that there are causes for reservation (in line with Bender et al.).

We disagree with the view that LLMs should be considered collaborators or research assistants (Schulz et al.). One can instruct students or research assistants, correct their mistakes, and anticipate that they will learn from them. One may also question their reasons or their reasoning and get answers and expect accountability. Finally, one may also get insight into their values and their motivations, and trust or distrust them accordingly. LLMs are not introspective, lack metacognition, and have no values, at least not in the way humans do. Indeed, our inability to understand why they make the errors they do or when they will make them impairs our ability to understand their limits, especially on the edges of knowledge, where their training corpus is arguably less robust. Moreover, although LLMs move from the same foundations of previous language models (Schulz et al.), they are significantly more opaque and complex. As a result, maintaining the ever-important scientific value of transparency can be challenging and necessitates further development of practices and strategies to ensure its preservation.

Nevertheless, we disagree that such concerns should prevent scientific applications of LLMs. It is unrealistic to presume that LLMs won’t be used because of the risks involved, and banning them could do more harm than good: given the current trend, if prohibited, they would likely be used covertly, exacerbating the already-worrying transparency issues. Certainly, we need to pursue a critical and not starry-eyed understanding of LLMs and maintain a clear-eyed assessment of the potential risks of use. However, there are ways of employing them that can improve the quality of science, as long as the researcher is kept at the center of the process. LLMs are tools and, as such, must be carefully evaluated in their applications. This applies to any tool, including the existing alternatives discussed by Bender et al., which, although optimized for specific scientific purposes, are not immune from mistakes and whose degree of reliability always needs careful scrutiny. At the end of the day, the responsibility falls upon the shoulders of the researchers who use the tools. It is, hence, crucial to establish principles and values that guide our decisions—whether one applies LLMs or any other method.

Ultimately, we mostly concur with Botvinick and Gershman: the impact of LLMs on the future practice of science cannot be fully predicted, but science is a humanistic and human enterprise and must remain so, motivating curbs to LLM use. Our perspective highlighted the normative aspects in terms of core values that should guide their use today, while Botvinick and Gershman seeks to identify the principles and values for the future, deciding what should remain exclusively human even when AI becomes fully capable of performing every step of scientific inquiry as well as upholding values such as accountability, transparency, and fairness. The two perspectives complement each other in stimulating discussions about what should guide the way we integrate AI into our scientific practices.

\section*{Response \textmd{by Matthew M. Botvinick and Samuel J. Gershman}}

We see significant common ground across the other perspectives. We will focus here on one issue that gets to the heart of our perspective. Schulz et al. characterize LLMs as closer to collaborators than to tools. This raises critical issues of accountability, as pointed out by Bender et al. and Marelli et al. Some of these issues are currently being grappled with, while others will become more salient in the future as the technology advances. In particular, accountability is a fundamentally human concept: humans are the only currently existing agents that are accountable in the sense that they have ultimate control over their own actions and voluntarily submit to a system that regulates these actions. Extending this concept to artificial agents would entail a profound shift in our attitudes, essentially requiring us to acknowledge the personhood of such agents.

This shift, if it ever happens, will have ramifications far beyond science. Policymakers are already starting to wrestle with the question of how accountability should operate in a world where AI systems are increasingly autonomous, and the issues can get quite complex. The difficulties can be bounded, however, in domains where humans are able to draw clear boundaries around what role they will permit AI systems to play. In science, we believe these boundaries should be firm and restrictive, limiting key decisions — and thus accountability — to human scientists.  

Ultimately, we are interested in the limit case where the limits imposed on AI are sociological, moral, and juristic, rather than technological. To regard LLMs as genuine collaborators rather than sophisticated tools, we would need to acknowledge attributes of personhood that go far beyond the mere practice of science. Our view is that AI, no matter how intelligent, should remain a tool, because ceding personhood to artificial agents would have undesirable consequences. It’s one thing for an AI scientist to tell us that there is a better way to fold proteins or design nuclear reactors, but it’s quite another thing for it to tell us that it would rather be studying some other problem. It would also be quite a shock to be told by an AI scientist that it’s solved an important problem but that it doesn’t feel like trying to explain it to a human. As we argued in our perspective, the choices of what to study and which explanations count are irreducibly human.

\section*{Conclusion}

We have presented four different perspectives centering around the question \say{how should the advent of LLMs affect the practice of science?} Schulz et al. argued that \say{working with LLMs will not be fundamentally different from working with other collaborators, such as research assistants or doctoral students.} Bender et al. described a suite of problems with using LLMs in scientific activity and argued that many uses of LLMs are \say{contrary to the norms of science.} Marelli et al. called for \say{clear principles guiding the way this technology should be used,} including transparency, accountability, and fairness. Finally, Botvinick and Gershman advocated that \say{two core aspects of scientific work should be reserved to human scientists}, namely deciding on what problems to work on and that human understanding remains the goal of science.

Yet, even though there was substantial disagreement, there were also important common themes. In particular, all parties emphasized the social nature of science and the importance of protecting scientific integrity and standards. In modern times, these core values are more important than ever before, and we\,---\,as a community\,---\,will have to continuously reevaluate how to protect them.

\bibliography{sample}

\begin{thebibliography}{10}
\urlstyle{rm}
\expandafter\ifx\csname url\endcsname\relax
  \def\url#1{\texttt{#1}}\fi
\expandafter\ifx\csname urlprefix\endcsname\relax\def\urlprefix{URL }\fi
\expandafter\ifx\csname doiprefix\endcsname\relax\def\doiprefix{DOI: }\fi
\providecommand{\bibinfo}[2]{#2}
\providecommand{\eprint}[2][]{\url{#2}}

\bibitem{bengio2000neural}
\bibinfo{author}{Bengio, Y.}, \bibinfo{author}{Ducharme, R.} \& \bibinfo{author}{Vincent, P.}
\newblock \bibinfo{journal}{\bibinfo{title}{A neural probabilistic language model}}.
\newblock {\emph{\JournalTitle{Advances in neural information processing systems}}} \textbf{\bibinfo{volume}{13}} (\bibinfo{year}{2000}).

\bibitem{Jurafsky2009}
\bibinfo{author}{Jurafsky, D.} \& \bibinfo{author}{Martin, J.~H.}
\newblock \emph{\bibinfo{title}{Speech and language processing : an introduction to natural language processing, computational linguistics, and speech recognition}} (\bibinfo{publisher}{Pearson Prentice Hall}, \bibinfo{year}{2009}).

\bibitem{brown2020language}
\bibinfo{author}{Brown, T.} \emph{et~al.}
\newblock \bibinfo{journal}{\bibinfo{title}{Language models are few-shot learners}}.
\newblock {\emph{\JournalTitle{Advances in neural information processing systems}}} \textbf{\bibinfo{volume}{33}}, \bibinfo{pages}{1877--1901} (\bibinfo{year}{2020}).

\bibitem{drori2022neural}
\bibinfo{author}{Drori, I.} \emph{et~al.}
\newblock \bibinfo{journal}{\bibinfo{title}{A neural network solves, explains, and generates university math problems by program synthesis and few-shot learning at human level}}.
\newblock {\emph{\JournalTitle{Proceedings of the National Academy of Sciences}}} \textbf{\bibinfo{volume}{119}}, \bibinfo{pages}{e2123433119} (\bibinfo{year}{2022}).

\bibitem{kocmi2023large}
\bibinfo{author}{Kocmi, T.} \& \bibinfo{author}{Federmann, C.}
\newblock \bibinfo{title}{Large language models are state-of-the-art evaluators of translation quality}.
\newblock In \bibinfo{editor}{Nurminen, M.} \emph{et~al.} (eds.) \emph{\bibinfo{booktitle}{Proceedings of the 24th Annual Conference of the European Association for Machine Translation}}, \bibinfo{pages}{193--203} (\bibinfo{publisher}{European Association for Machine Translation}, \bibinfo{address}{Tampere, Finland}, \bibinfo{year}{2023}).

\bibitem{katz2023gpt}
\bibinfo{author}{Katz, D.~M.}, \bibinfo{author}{Bommarito, M.~J.}, \bibinfo{author}{Gao, S.} \& \bibinfo{author}{Arredondo, P.}
\newblock \bibinfo{journal}{\bibinfo{title}{Gpt-4 passes the bar exam}}.
\newblock {\emph{\JournalTitle{Available at SSRN 4389233}}}  (\bibinfo{year}{2023}).

\bibitem{eloundou2023gpts}
\bibinfo{author}{Eloundou, T.}, \bibinfo{author}{Manning, S.}, \bibinfo{author}{Mishkin, P.} \& \bibinfo{author}{Rock, D.}
\newblock \bibinfo{journal}{\bibinfo{title}{Gpts are gpts: An early look at the labor market impact potential of large language models}}.
\newblock {\emph{\JournalTitle{arXiv:2303.10130. Unpublished preprint}}}  (\bibinfo{year}{2023}).

\bibitem{kasneci2023chatgpt}
\bibinfo{author}{Kasneci, E.} \emph{et~al.}
\newblock \bibinfo{journal}{\bibinfo{title}{Chatgpt for good? on opportunities and challenges of large language models for education}}.
\newblock {\emph{\JournalTitle{Learning and Individual Differences}}} \textbf{\bibinfo{volume}{103}}, \bibinfo{pages}{102274} (\bibinfo{year}{2023}).

\bibitem{peres2023chatgpt}
\bibinfo{author}{Peres, R.}, \bibinfo{author}{Schreier, M.}, \bibinfo{author}{Schweidel, D.} \& \bibinfo{author}{Sorescu, A.}
\newblock \bibinfo{journal}{\bibinfo{title}{On chatgpt and beyond: How generative artificial intelligence may affect research, teaching, and practice}}.
\newblock {\emph{\JournalTitle{International Journal of Research in Marketing}}}  (\bibinfo{year}{2023}).

\bibitem{lund2023chatting}
\bibinfo{author}{Lund, B.~D.} \& \bibinfo{author}{Wang, T.}
\newblock \bibinfo{journal}{\bibinfo{title}{Chatting about chatgpt: how may ai and gpt impact academia and libraries?}}
\newblock {\emph{\JournalTitle{Library Hi Tech News}}} \textbf{\bibinfo{volume}{40}}, \bibinfo{pages}{26--29} (\bibinfo{year}{2023}).

\bibitem{hill2023chat}
\bibinfo{author}{Hill-Yardin, E.~L.}, \bibinfo{author}{Hutchinson, M.~R.}, \bibinfo{author}{Laycock, R.} \& \bibinfo{author}{Spencer, S.~J.}
\newblock \bibinfo{journal}{\bibinfo{title}{A chat (gpt) about the future of scientific publishing}}.
\newblock {\emph{\JournalTitle{Brain Behav Immun}}} \textbf{\bibinfo{volume}{110}}, \bibinfo{pages}{152--154} (\bibinfo{year}{2023}).

\bibitem{zheng2023chatgpt}
\bibinfo{author}{Zheng, H.} \& \bibinfo{author}{Zhan, H.}
\newblock \bibinfo{journal}{\bibinfo{title}{Chatgpt in scientific writing: a cautionary tale}}.
\newblock {\emph{\JournalTitle{The American Journal of Medicine}}}  (\bibinfo{year}{2023}).

\bibitem{lund2023chatgpt}
\bibinfo{author}{Lund, B.~D.} \emph{et~al.}
\newblock \bibinfo{journal}{\bibinfo{title}{Chatgpt and a new academic reality: Artificial intelligence-written research papers and the ethics of the large language models in scholarly publishing}}.
\newblock {\emph{\JournalTitle{Journal of the Association for Information Science and Technology}}} \textbf{\bibinfo{volume}{74}}, \bibinfo{pages}{570--581} (\bibinfo{year}{2023}).

\bibitem{transformer2022can}
\bibinfo{author}{Transformer, G. G.~P.}, \bibinfo{author}{Thunstr{\"o}m, A.~O.} \& \bibinfo{author}{Steingrimsson, S.}
\newblock \bibinfo{journal}{\bibinfo{title}{Can gpt-3 write an academic paper on itself, with minimal human input?}}
\newblock {\emph{\JournalTitle{Unpublished}}}  (\bibinfo{year}{2022}).

\bibitem{birhane2023science}
\bibinfo{author}{Birhane, A.}, \bibinfo{author}{Kasirzadeh, A.}, \bibinfo{author}{Leslie, D.} \& \bibinfo{author}{Wachter, S.}
\newblock \bibinfo{journal}{\bibinfo{title}{Science in the age of large language models}}.
\newblock {\emph{\JournalTitle{Nature Reviews Physics}}} \bibinfo{pages}{1--4} (\bibinfo{year}{2023}).

\bibitem{fecher2023friend}
\bibinfo{author}{Fecher, B.}, \bibinfo{author}{Hebing, M.}, \bibinfo{author}{Laufer, M.}, \bibinfo{author}{Pohle, J.} \& \bibinfo{author}{Sofsky, F.}
\newblock \bibinfo{journal}{\bibinfo{title}{Friend or foe? exploring the implications of large language models on the science system}}.
\newblock {\emph{\JournalTitle{arXiv:2306.09928. Unpublished preprint}}}  (\bibinfo{year}{2023}).

\bibitem{stokelwalker2023chatgpt}
\bibinfo{author}{Stokel-Walker, C.} \& \bibinfo{author}{Van~Noorden, R.}
\newblock \bibinfo{title}{What chatgpt and generative ai mean for science}, \doiprefix\url{10.1038/d41586-023-00340-6} (\bibinfo{year}{2023}).

\bibitem{taylor2022galactica}
\bibinfo{author}{Taylor, R.} \emph{et~al.}
\newblock \bibinfo{journal}{\bibinfo{title}{Galactica: A large language model for science}}.
\newblock {\emph{\JournalTitle{arXiv:2211.09085. Unpublished preprint}}}  (\bibinfo{year}{2022}).

\bibitem{tao2023}
\bibinfo{title}{Embracing change and resetting expectations}.
\newblock \bibinfo{howpublished}{\url{https://unlocked.microsoft.com/ai-anthology/terence-tao/}}.
\newblock \bibinfo{note}{Accessed: 2023-09-04}.

\bibitem{heaven2022why}
\bibinfo{author}{Heaven, W.~D.}
\newblock \bibinfo{title}{Why {Meta}’s latest large language model survived only three days online} (\bibinfo{year}{2022}).

\bibitem{bender2021dangers}
\bibinfo{author}{Bender, E.~M.}, \bibinfo{author}{Gebru, T.}, \bibinfo{author}{McMillan-Major, A.} \& \bibinfo{author}{Shmitchell, S.}
\newblock \bibinfo{title}{On the dangers of stochastic parrots: Can language models be too big?}
\newblock In \emph{\bibinfo{booktitle}{Proceedings of the 2021 ACM conference on fairness, accountability, and transparency}}, \bibinfo{pages}{610--623} (\bibinfo{year}{2021}).

\bibitem{arkoudas2023gpt}
\bibinfo{author}{Arkoudas, K.}
\newblock \bibinfo{journal}{\bibinfo{title}{Gpt-4 can't reason}}.
\newblock {\emph{\JournalTitle{arXiv:2308.03762. Unpublished preprint}}}  (\bibinfo{year}{2023}).

\bibitem{gilardi2023chatgpt}
\bibinfo{author}{Gilardi, F.}, \bibinfo{author}{Alizadeh, M.} \& \bibinfo{author}{Kubli, M.}
\newblock \bibinfo{journal}{\bibinfo{title}{Chatgpt outperforms crowd workers for text-annotation tasks}}.
\newblock {\emph{\JournalTitle{Proceedings of the National Academy of Sciences}}} \textbf{\bibinfo{volume}{120}}, \bibinfo{pages}{e2305016120}, \doiprefix\url{10.1073/pnas.2305016120} (\bibinfo{year}{2023}).
\newblock \eprint{https://www.pnas.org/doi/pdf/10.1073/pnas.2305016120}.

\bibitem{wulff2023automated}
\bibinfo{author}{Wulff, D.~U.} \& \bibinfo{author}{Mata, R.}
\newblock \bibinfo{journal}{\bibinfo{title}{Automated jingle--jangle detection: Using embeddings to tackle taxonomic incommensurability}}.
\newblock {\emph{\JournalTitle{PsyArXiv}}} \doiprefix\url{https://doi.org/10.31234/osf.io/9h7aw} (\bibinfo{year}{2023}).

\bibitem{dillion2023can}
\bibinfo{author}{Dillion, D.}, \bibinfo{author}{Tandon, N.}, \bibinfo{author}{Gu, Y.} \& \bibinfo{author}{Gray, K.}
\newblock \bibinfo{journal}{\bibinfo{title}{Can ai language models replace human participants?}}
\newblock {\emph{\JournalTitle{Trends in Cognitive Sciences}}}  (\bibinfo{year}{2023}).

\bibitem{hutson2023pigbots}
\bibinfo{author}{Hutson, M.}
\newblock \bibinfo{journal}{\bibinfo{title}{Guinea pigbots}}.
\newblock {\emph{\JournalTitle{Science (New York, N.Y.)}}} \textbf{\bibinfo{volume}{381}}, \bibinfo{pages}{121--123}, \doiprefix\url{10.1126/science.adj6791} (\bibinfo{year}{2023}).

\bibitem{roziere2023codellama}
\bibinfo{author}{Rozi{\`{e}}re, B.} \emph{et~al.}
\newblock \bibinfo{journal}{\bibinfo{title}{Code llama: Open foundation models for code}}.
\newblock {\emph{\JournalTitle{arXiv:2308.12950. Unpublished preprint}}}  (\bibinfo{year}{2023}).

\bibitem{sanmarchi2023step}
\bibinfo{author}{Sanmarchi, F.} \emph{et~al.}
\newblock \bibinfo{journal}{\bibinfo{title}{A step-by-step researcher's guide to the use of an ai-based transformer in epidemiology: an exploratory analysis of chatgpt using the strobe checklist for observational studies}}.
\newblock {\emph{\JournalTitle{Journal of Public Health}}} \bibinfo{pages}{1--36} (\bibinfo{year}{2023}).

\bibitem{dehouche2021plagiarism}
\bibinfo{author}{Dehouche, N.}
\newblock \bibinfo{journal}{\bibinfo{title}{Plagiarism in the age of massive generative pre-trained transformers (gpt-3)}}.
\newblock {\emph{\JournalTitle{Ethics in Science and Environmental Politics}}} \textbf{\bibinfo{volume}{21}}, \bibinfo{pages}{17--23} (\bibinfo{year}{2021}).

\bibitem{liang2021towards}
\bibinfo{author}{Liang, P.~P.}, \bibinfo{author}{Wu, C.}, \bibinfo{author}{Morency, L.-P.} \& \bibinfo{author}{Salakhutdinov, R.}
\newblock \bibinfo{title}{Towards understanding and mitigating social biases in language models}.
\newblock In \emph{\bibinfo{booktitle}{International Conference on Machine Learning}}, \bibinfo{pages}{6565--6576} (\bibinfo{organization}{PMLR}, \bibinfo{year}{2021}).

\bibitem{coda2023inducing}
\bibinfo{author}{Coda-Forno, J.} \emph{et~al.}
\newblock \bibinfo{journal}{\bibinfo{title}{Inducing anxiety in large language models increases exploration and bias}}.
\newblock {\emph{\JournalTitle{arXiv:2304.11111. Unpublished preprint}}}  (\bibinfo{year}{2023}).

\bibitem{hutchinson2020socialbias}
\bibinfo{author}{Hutchinson, B.} \emph{et~al.}
\newblock \bibinfo{title}{Social biases in {NLP} models as barriers for persons with disabilities}.
\newblock In \emph{\bibinfo{booktitle}{Proceedings of the 58th Annual Meeting of the Association for Computational Linguistics}}, \bibinfo{pages}{5491--5501}, \doiprefix\url{10.18653/v1/2020.acl-main.487} (\bibinfo{publisher}{Association for Computational Linguistics}, \bibinfo{year}{2020}).

\bibitem{carlini2021extracting}
\bibinfo{author}{Carlini, N.} \emph{et~al.}
\newblock \bibinfo{title}{Extracting training data from large language models}.
\newblock In \emph{\bibinfo{booktitle}{30th USENIX Security Symposium (USENIX Security 21)}}, \bibinfo{pages}{2633--2650} (\bibinfo{year}{2021}).

\bibitem{king2023place}
\bibinfo{author}{King, M.~R.}
\newblock \bibinfo{journal}{\bibinfo{title}{A place for large language models in scientific publishing, apart from credited authorship}}.
\newblock {\emph{\JournalTitle{Cellular and Molecular Bioengineering}}} \bibinfo{pages}{1--4} (\bibinfo{year}{2023}).

\bibitem{herbold2023ai}
\bibinfo{author}{Herbold, S.}, \bibinfo{author}{Hautli-Janisz, A.}, \bibinfo{author}{Heuer, U.}, \bibinfo{author}{Kikteva, Z.} \& \bibinfo{author}{Trautsch, A.}
\newblock \bibinfo{journal}{\bibinfo{title}{Ai, write an essay for me: A large-scale comparison of human-written versus chatgpt-generated essays}}.
\newblock {\emph{\JournalTitle{arXiv:2304.14276. Unpublished preprint}}}  (\bibinfo{year}{2023}).

\bibitem{poldrack2023ai}
\bibinfo{author}{Poldrack, R.~A.}, \bibinfo{author}{Lu, T.} \& \bibinfo{author}{Begu{\v{s}}, G.}
\newblock \bibinfo{journal}{\bibinfo{title}{Ai-assisted coding: Experiments with gpt-4}}.
\newblock {\emph{\JournalTitle{arXiv:2304.13187. Unpublished preprint}}}  (\bibinfo{year}{2023}).

\bibitem{goyal2022news}
\bibinfo{author}{Goyal, T.}, \bibinfo{author}{Li, J.~J.} \& \bibinfo{author}{Durrett, G.}
\newblock \bibinfo{journal}{\bibinfo{title}{News summarization and evaluation in the era of gpt-3}}.
\newblock {\emph{\JournalTitle{arXiv:2209.12356. Unpublished preprint}}}  (\bibinfo{year}{2022}).

\bibitem{laion2023}
\bibinfo{title}{Towards a transparent ai future: The call for less regulatory hurdles on open-source ai in europe}.
\newblock \bibinfo{howpublished}{\url{https://laion.ai/blog/transparent-ai/}}.
\newblock \bibinfo{note}{Accessed: 2023-10-22}.

\bibitem{liang2023can}
\bibinfo{author}{Liang, W.} \emph{et~al.}
\newblock \bibinfo{journal}{\bibinfo{title}{Can large language models provide useful feedback on research papers? a large-scale empirical analysis}}.
\newblock {\emph{\JournalTitle{arXiv:2310.01783. Unpublished preprint}}}  (\bibinfo{year}{2023}).

\bibitem{Ben:Kol:20}
\bibinfo{author}{Bender, E.~M.} \& \bibinfo{author}{Koller, A.}
\newblock \bibinfo{title}{Climbing towards {NLU}: {On} meaning, form, and understanding in the age of data}.
\newblock In \emph{\bibinfo{booktitle}{Proceedings of the 58th Annual Meeting of the Association for Computational Linguistics}}, \bibinfo{pages}{5185--5198}, \doiprefix\url{10.18653/v1/2020.acl-main.463} (\bibinfo{publisher}{Association for Computational Linguistics}, \bibinfo{address}{Online}, \bibinfo{year}{2020}).

\bibitem{Wan:Wan:Lv:19}
\bibinfo{author}{Wang, D.}, \bibinfo{author}{Wang, X.} \& \bibinfo{author}{Lv, S.}
\newblock \bibinfo{journal}{\bibinfo{title}{An overview of end-to-end automatic speech recognition}}.
\newblock {\emph{\JournalTitle{Symmetry}}} \textbf{\bibinfo{volume}{11}}, \doiprefix\url{10.3390/sym11081018} (\bibinfo{year}{2019}).

\bibitem{Har:Moo:17}
\bibinfo{author}{Hartsuiker, R.~J.} \& \bibinfo{author}{Moors, A.}
\newblock \bibinfo{title}{On the automaticity of language processing}.
\newblock In \bibinfo{editor}{Schmid, H.-J.} (ed.) \emph{\bibinfo{booktitle}{Entrenchment and the Psychology of Language Learning: {How} We Reorganize and Adapt Linguistic Knowledge}}, \doiprefix\url{https://doi-org.offcampus.lib.washington.edu/10.1037/15969-010} (\bibinfo{publisher}{American Psychological Association; De Gruyter Mouton}, \bibinfo{year}{2017}).

\bibitem{Kin:Ana:Aut:23}
\bibinfo{author}{Kinney, R.~M.} \emph{et~al.}
\newblock \bibinfo{journal}{\bibinfo{title}{The semantic scholar open data platform}}.
\newblock {\emph{\JournalTitle{ArXiv. Unpublished preprint}}} \textbf{\bibinfo{volume}{abs/2301.10140}} (\bibinfo{year}{2023}).

\bibitem{narayan2018ranking}
\bibinfo{author}{Narayan, S.}, \bibinfo{author}{Cohen, S.~B.} \& \bibinfo{author}{Lapata, M.}
\newblock \bibinfo{title}{Ranking sentences for extractive summarization with reinforcement learning}.
\newblock In \bibinfo{editor}{Walker, M.}, \bibinfo{editor}{Ji, H.} \& \bibinfo{editor}{Stent, A.} (eds.) \emph{\bibinfo{booktitle}{Proceedings of the 2018 Conference of the North {A}merican Chapter of the Association for Computational Linguistics: Human Language Technologies, Volume 1 (Long Papers)}}, \bibinfo{pages}{1747--1759}, \doiprefix\url{10.18653/v1/N18-1158} (\bibinfo{publisher}{Association for Computational Linguistics}, \bibinfo{address}{New Orleans, Louisiana}, \bibinfo{year}{2018}).

\bibitem{hodel2023response}
\bibinfo{author}{Hodel, D.} \& \bibinfo{author}{West, J.}
\newblock \bibinfo{title}{Response: Emergent analogical reasoning in large language models} (\bibinfo{year}{2023}).
\newblock \eprint{2308.16118}.

\bibitem{törnberg2023simulating}
\bibinfo{author}{Törnberg, P.}, \bibinfo{author}{Valeeva, D.}, \bibinfo{author}{Uitermark, J.} \& \bibinfo{author}{Bail, C.}
\newblock \bibinfo{title}{Simulating social media using large language models to evaluate alternative news feed algorithms} (\bibinfo{year}{2023}).
\newblock \eprint{2310.05984}.

\bibitem{arg:us:ful:2023}
\bibinfo{author}{Argyle, L.~P.} \emph{et~al.}
\newblock \bibinfo{journal}{\bibinfo{title}{Out of one, many: Using language models to simulate human samples}}.
\newblock {\emph{\JournalTitle{Political Analysis}}} \textbf{\bibinfo{volume}{31}}, \bibinfo{pages}{337–351}, \doiprefix\url{10.1017/pan.2023.2} (\bibinfo{year}{2023}).

\bibitem{Gil:Ali:Kub:23}
\bibinfo{author}{Gilardi, F.}, \bibinfo{author}{Alizadeh, M.} \& \bibinfo{author}{Kubli, M.}
\newblock \bibinfo{journal}{\bibinfo{title}{Chatgpt outperforms crowd workers for text-annotation tasks}}.
\newblock {\emph{\JournalTitle{Proceedings of the National Academy of Sciences}}} \textbf{\bibinfo{volume}{120}}, \bibinfo{pages}{e2305016120}, \doiprefix\url{10.1073/pnas.2305016120} (\bibinfo{year}{2023}).
\newblock \eprint{https://www.pnas.org/doi/pdf/10.1073/pnas.2305016120}.

\bibitem{conroy2023chatgpt}
\bibinfo{author}{Conroy, G.}
\newblock \bibinfo{journal}{\bibinfo{title}{How {ChatGPT} and other {AI} tools could disrupt scientific publishing}}.
\newblock {\emph{\JournalTitle{Nature}}} \textbf{\bibinfo{volume}{622}}, \bibinfo{pages}{234--236} (\bibinfo{year}{2023}).

\bibitem{latour2013laboratory}
\bibinfo{author}{Latour, B.} \& \bibinfo{author}{Woolgar, S.}
\newblock \emph{\bibinfo{title}{Laboratory life: The construction of scientific facts}} (\bibinfo{publisher}{Princeton university press}, \bibinfo{year}{2013}).

\bibitem{partha1994toward}
\bibinfo{author}{Partha, D.} \& \bibinfo{author}{David, P.~A.}
\newblock \bibinfo{journal}{\bibinfo{title}{Toward a new economics of science}}.
\newblock {\emph{\JournalTitle{Research policy}}} \textbf{\bibinfo{volume}{23}}, \bibinfo{pages}{487--521} (\bibinfo{year}{1994}).

\bibitem{amano2023manifold}
\bibinfo{author}{Amano, T.} \emph{et~al.}
\newblock \bibinfo{journal}{\bibinfo{title}{The manifold costs of being a non-native english speaker in science}}.
\newblock {\emph{\JournalTitle{PLoS Biology}}} \textbf{\bibinfo{volume}{21}}, \bibinfo{pages}{e3002184} (\bibinfo{year}{2023}).

\bibitem{jumper2021highly}
\bibinfo{author}{Jumper, J.} \emph{et~al.}
\newblock \bibinfo{journal}{\bibinfo{title}{Highly accurate protein structure prediction with alphafold}}.
\newblock {\emph{\JournalTitle{Nature}}} \textbf{\bibinfo{volume}{596}}, \bibinfo{pages}{583--589} (\bibinfo{year}{2021}).

\bibitem{gunther2019vector}
\bibinfo{author}{G{\"u}nther, F.}, \bibinfo{author}{Rinaldi, L.} \& \bibinfo{author}{Marelli, M.}
\newblock \bibinfo{journal}{\bibinfo{title}{Vector-space models of semantic representation from a cognitive perspective: A discussion of common misconceptions}}.
\newblock {\emph{\JournalTitle{Perspectives on Psychological Science}}} \textbf{\bibinfo{volume}{14}}, \bibinfo{pages}{1006--1033} (\bibinfo{year}{2019}).

\bibitem{golchin2023time}
\bibinfo{author}{Golchin, S.} \& \bibinfo{author}{Surdeanu, M.}
\newblock \bibinfo{journal}{\bibinfo{title}{Time travel in llms: Tracing data contamination in large language models}}.
\newblock {\emph{\JournalTitle{arXiv:2308.08493. Unpublished preprint}}}  (\bibinfo{year}{2023}).

\bibitem{li2023starcoder}
\bibinfo{author}{Li, R.} \emph{et~al.}
\newblock \bibinfo{journal}{\bibinfo{title}{Starcoder: may the source be with you!}}
\newblock {\emph{\JournalTitle{arXiv:2305.06161. Unpublished preprint}}}  (\bibinfo{year}{2023}).

\bibitem{walters2023fabrication}
\bibinfo{author}{Walters, W.~H.} \& \bibinfo{author}{Wilder, E.~I.}
\newblock \bibinfo{journal}{\bibinfo{title}{Fabrication and errors in the bibliographic citations generated by chatgpt}}.
\newblock {\emph{\JournalTitle{Scientific Reports}}} \textbf{\bibinfo{volume}{13}}, \bibinfo{pages}{14045} (\bibinfo{year}{2023}).

\bibitem{kocon2023chatgpt}
\bibinfo{author}{Koco{\'n}, J.} \emph{et~al.}
\newblock \bibinfo{journal}{\bibinfo{title}{Chatgpt: Jack of all trades, master of none}}.
\newblock {\emph{\JournalTitle{Information Fusion}}} \bibinfo{pages}{101861} (\bibinfo{year}{2023}).

\bibitem{liu2023evaluating}
\bibinfo{author}{Liu, H.} \emph{et~al.}
\newblock \bibinfo{journal}{\bibinfo{title}{Evaluating the logical reasoning ability of chatgpt and gpt-4}}.
\newblock {\emph{\JournalTitle{arXiv:2304.03439. Unpublished preprint}}}  (\bibinfo{year}{2023}).

\bibitem{durmus2023towards}
\bibinfo{author}{Durmus, E.} \emph{et~al.}
\newblock \bibinfo{journal}{\bibinfo{title}{Towards measuring the representation of subjective global opinions in language models}}.
\newblock {\emph{\JournalTitle{arXiv:2306.16388. Unpublished preprint}}}  (\bibinfo{year}{2023}).

\bibitem{atari2023humans}
\bibinfo{author}{Atari, M.}, \bibinfo{author}{Xue, M.~J.}, \bibinfo{author}{Park, P.~S.}, \bibinfo{author}{Blasi, D.} \& \bibinfo{author}{Henrich, J.}
\newblock \bibinfo{journal}{\bibinfo{title}{Which humans?}}
\newblock {\emph{\JournalTitle{PsyArXiv. Unpublished preprint}}}  (\bibinfo{year}{2023}).

\bibitem{santurkar2023whose}
\bibinfo{author}{Santurkar, S.} \emph{et~al.}
\newblock \bibinfo{journal}{\bibinfo{title}{Whose opinions do language models reflect?}}
\newblock {\emph{\JournalTitle{arXiv:2303.17548. Unpublished preprint}}}  (\bibinfo{year}{2023}).

\bibitem{flaherty2023}
\bibinfo{author}{Flaherty, C.}
\newblock \bibinfo{title}{The peer review crisis}.
\newblock \bibinfo{howpublished}{\url{https://www.insidehighered.com/news/2022/06/13/peer-review-crisis-creates-problems-journals-and-scholars}}.
\newblock \bibinfo{note}{Accessed: 2023-10-30}.

\bibitem{park2023papers}
\bibinfo{author}{Park, M.}, \bibinfo{author}{Leahey, E.} \& \bibinfo{author}{Funk, R.~J.}
\newblock \bibinfo{journal}{\bibinfo{title}{Papers and patents are becoming less disruptive over time}}.
\newblock {\emph{\JournalTitle{Nature}}} \textbf{\bibinfo{volume}{613}}, \bibinfo{pages}{138--144} (\bibinfo{year}{2023}).

\bibitem{davies2021advancing}
\bibinfo{author}{Davies, A.} \emph{et~al.}
\newblock \bibinfo{journal}{\bibinfo{title}{Advancing mathematics by guiding human intuition with ai}}.
\newblock {\emph{\JournalTitle{Nature}}} \textbf{\bibinfo{volume}{600}}, \bibinfo{pages}{70--74} (\bibinfo{year}{2021}).

\bibitem{gabriel2021challenge}
\bibinfo{author}{Gabriel, I.} \& \bibinfo{author}{Ghazavi, V.}
\newblock \bibinfo{title}{{The Challenge of Value Alignment: From Fairer Algorithms to AI Safety}}.
\newblock In \emph{\bibinfo{booktitle}{{The Oxford Handbook of Digital Ethics}}}, \doiprefix\url{10.1093/oxfordhb/9780198857815.013.18} (\bibinfo{publisher}{Oxford University Press}).
\newblock \eprint{https://academic.oup.com/book/0/chapter/337809435/chapter-ag-pdf/50148600/book\_37078\_section\_337809435.ag.pdf}.

\bibitem{silver2017mastering}
\bibinfo{author}{Silver, D.} \emph{et~al.}
\newblock \bibinfo{journal}{\bibinfo{title}{Mastering the game of go without human knowledge}}.
\newblock {\emph{\JournalTitle{nature}}} \textbf{\bibinfo{volume}{550}}, \bibinfo{pages}{354--359} (\bibinfo{year}{2017}).

\bibitem{lemos2023rediscovering}
\bibinfo{author}{Lemos, P.}, \bibinfo{author}{Jeffrey, N.}, \bibinfo{author}{Cranmer, M.}, \bibinfo{author}{Ho, S.} \& \bibinfo{author}{Battaglia, P.}
\newblock \bibinfo{journal}{\bibinfo{title}{Rediscovering orbital mechanics with machine learning}}.
\newblock {\emph{\JournalTitle{Machine Learning: Science and Technology}}} \textbf{\bibinfo{volume}{4}}, \bibinfo{pages}{045002} (\bibinfo{year}{2023}).

\bibitem{botvinick2023}
\bibinfo{author}{Botvinick, M.}
\newblock \bibinfo{title}{Have we lost our minds?}
\newblock \bibinfo{howpublished}{\url{https://medium.com/@matthew.botvinick/have-we-lost-our-minds-86d9125bd803}}.
\newblock \bibinfo{note}{Accessed: 2023-10-30}.

\bibitem{stiennon2020learning}
\bibinfo{author}{Stiennon, N.} \emph{et~al.}
\newblock \bibinfo{journal}{\bibinfo{title}{Learning to summarize with human feedback}}.
\newblock {\emph{\JournalTitle{Advances in Neural Information Processing Systems}}} \textbf{\bibinfo{volume}{33}}, \bibinfo{pages}{3008--3021} (\bibinfo{year}{2020}).

\bibitem{10.5555/3618408.3619349}
\bibinfo{author}{Longpre, S.} \emph{et~al.}
\newblock \bibinfo{title}{The flan collection: Designing data and methods for effective instruction tuning}.
\newblock In \emph{\bibinfo{booktitle}{Proceedings of the 40th International Conference on Machine Learning}}, ICML'23 (\bibinfo{publisher}{JMLR.org}, \bibinfo{year}{2023}).

\bibitem{taylor_principles_1913}
\bibinfo{author}{Taylor, F.~W.}
\newblock \emph{\bibinfo{title}{The {Principles} of {Scientific} {Management}}} (\bibinfo{publisher}{Harper}, \bibinfo{year}{1913}).

\end{thebibliography}

\section*{Acknowledgements}

This work has been partially funded by the ERC (853489 - DEXIM; 101087053 - BraveNewWord), by the DFG (2064/1 – Project number 390727645), the BMBF (Tübingen AI Center, FKZ: 01IS18039A) and as part of the Excellence Strategy of the German Federal and State Governments.

\section*{Author contributions statement}

\textbf{Project administration:} Marcel Binz, Stephan Alaniz \\
\textbf{Project supervision:} Zeynep Akata, Eric Schulz \\
\textbf{Perspective leaders:} Emily M.\ Bender, Marco Marelli, Matthew M. Botvinick, Eric Schulz \\
\textbf{Perspectives/responses - original draft:} Adina Roskies, Balazs Aczel, Carl T. Bergstrom, Emily M.\ Bender, Eric Schulz, Jevin West, Marco Marelli, Matthew M. Botvinick, Qiong Zhang \\
\textbf{Perspectives/responses - review \& editing:} all authors \\
\textbf{Introduction and conclusion - original draft:} Marcel Binz, Stephan Alaniz \\
\textbf{Introduction and conclusion - review \& editing:} Marcel Binz, Stephan Alaniz, Zeynep Akata, Eric Schulz

\end{document}